\documentclass[letterpaper]{article} 
\pdfoutput=1
\usepackage{aaai23}  
\usepackage{times}  
\usepackage{helvet}  
\usepackage{courier}  
\pdfoutput=1
\usepackage[hyphens]{url}  
\usepackage{graphicx} 
\usepackage{natbib}  
\usepackage{caption} 
\usepackage{color}
\usepackage{amsmath,epsfig}
\usepackage{makecell}
\usepackage[ruled, lined, linesnumbered, commentsnumbered, longend]{algorithm2e}

\usepackage{amssymb}
\usepackage{algorithmic}
\usepackage{booktabs}
\usepackage{multirow}
\usepackage{amsmath}
\usepackage{float}
\usepackage{comment}
\usepackage{bibentry}

\usepackage{newfloat}
\usepackage{listings}

\DeclareCaptionStyle{ruled}{labelfont=normalfont,labelsep=colon,strut=off} 
\floatstyle{ruled}
\newfloat{listing}{tb}{lst}{}
\floatname{listing}{Listing}
\title{SRoUDA: Meta Self-training for Robust Unsupervised Domain Adaptation}
\author{
    Wanqing Zhu\textsuperscript{\rm 1,2},
    Jia-Li Yin\textsuperscript{\rm 1,2},
    Bo-Hao Chen\textsuperscript{\rm 3\thanks{Corresponding author.}},
    Ximeng Liu\textsuperscript{\rm 1,2\footnotemark[1]}
}
\affiliations{
    \textsuperscript{\rm 1} Fujian Province Key Laboratory of Information Security of Network System, Fuzhou 350108, China\\
    \textsuperscript{\rm 2} College of Computer Science and Big Data, Fuzhou University, Fuzhou 350108, China\\
    \textsuperscript{\rm 3} Department of Computer Science and Engineering, Yuan Ze University, Taiwan\\
    
    wqingzhu00@163.com, jlyin@fzu.edu, bhchen@saturn.yzu.edu.tw, snbnix@gmail.com

}

\usepackage{bibentry}
\usepackage[colorlinks,linkcolor=red]{hyperref}
\begin{document}

\maketitle
\begin{abstract}
As acquiring manual labels on data could be costly, unsupervised domain adaptation (UDA), which transfers knowledge learned from a rich-label dataset to the unlabeled target dataset, is gaining increasing popularity. While extensive studies have been devoted to improving the model accuracy on target domain, an important issue of model \textit{robustness} is neglected. To make things worse, conventional adversarial training (AT) methods for improving model robustness are inapplicable under UDA scenario since they train models on adversarial examples that are generated by supervised loss function. In this paper, we present a new meta self-training pipeline, named SRoUDA, for improving adversarial robustness of UDA models. Based on self-training paradigm, SRoUDA starts with pre-training a source model by applying UDA baseline on source labeled data and taraget unlabeled data with a developed random masked augmentation (RMA), and then alternates between adversarial target model training on pseudo-labeled target data and fine-tuning source model by a meta step. While self-training allows the direct incorporation of AT in UDA, the meta step in SRoUDA further helps in mitigating error propagation from noisy pseudo labels. Extensive experiments on various benchmark datasets demonstrate the state-of-the-art performance of SRoUDA where it achieves significant model robustness improvement without harming clean accuracy.
Code is available at \url{https://github.com/VisionFlow/SRoUDA_main}

\end{abstract}

\section{Introduction}
\label{sec:intro}
\begin{figure}[t]
\begin{center}
\includegraphics[width=1.0\linewidth]{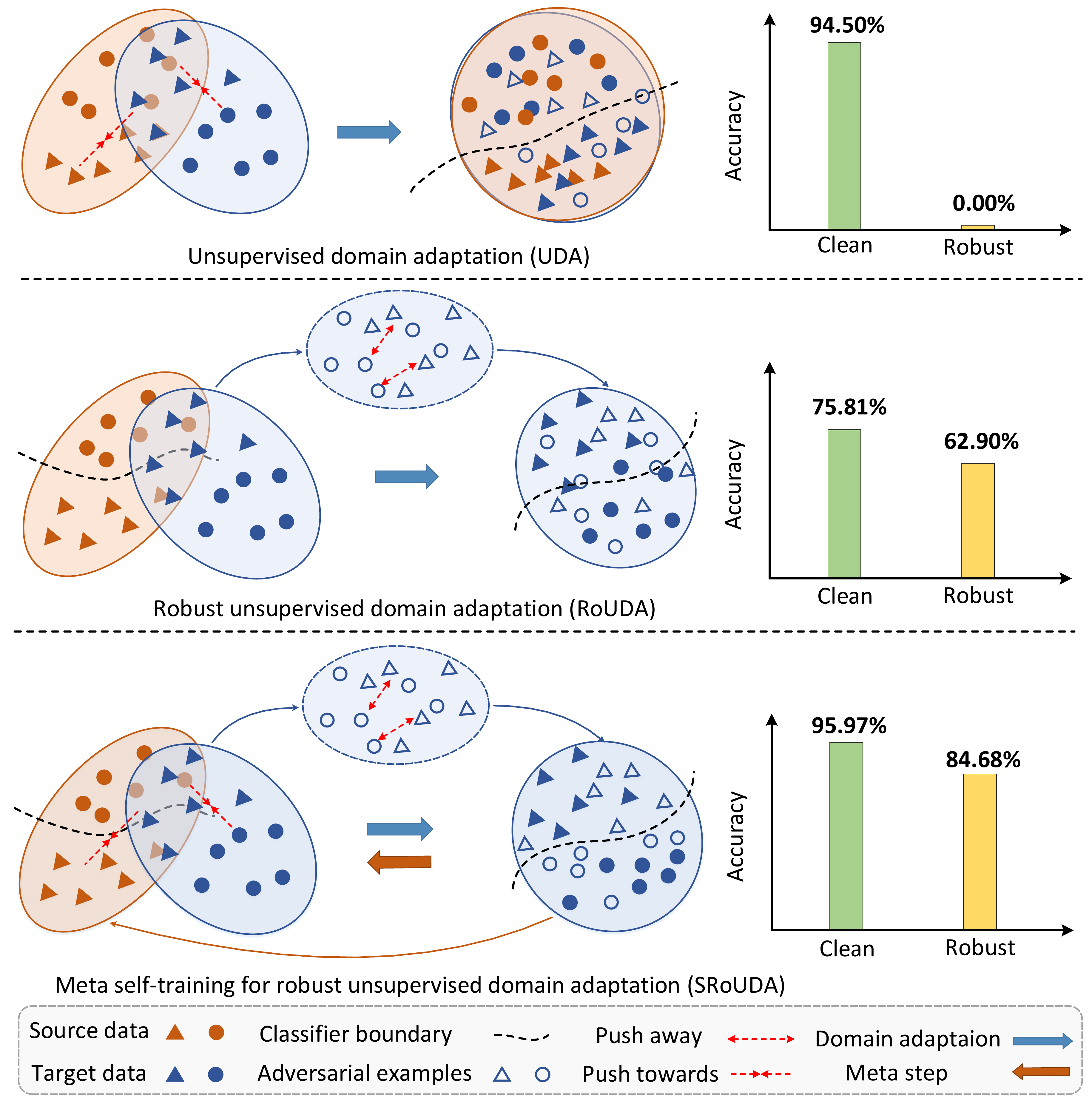}
\end{center}
\caption{Overview of different UDA schemes and their performance. Upper row: Conventional UDA. Middle row: Naive self-training for injecting AT into UDA. Bottom row: Our proposed SRoUDA. The statics are tested on ResNet-50 backbone on \textbf{A} $\xrightarrow{}$ \textbf{W} task in \textbf{Office-31} dataset. We use MDD~\cite{Zhang2019ICML} as the UDA baseline, and PGD-20 for evaluating model robustness.}
\label{fig:introduction}
\end{figure}

Deep neural networks (DNNs) have achieved impressive advances across a variety of machine learning tasks. However, these leaps come only when sufficient and well-labeled training data is available. For various label-scarce real-world scenarios, Unsupervised Domain Adaptation (UDA)~\cite{long2015ICML, Zhang2019ICML, Park_2022_CVPR, Fan_2022_CVPR} is widely studied and used. This process starts with a label-rich source dataset, and then transfers knowledge learned from the source domain to the target domain, usually by minimizing the distribution discrepancy between source and target domains.

While UDA approaches greatly reduce data requirements and are more practical for real-world scenarios, it neglects an important issue of model \textit{robustness}. Recent studies~\cite{Szegedy,madry2018towards,sehwag2022robust} have revealed that the DNN models are vulnerable to adversarial attacks, i.e., maliciously hand-crafted images which are similar looking to the original images but can lead to dramatic changes in model predictive behavior. For instance, given the UDA task $\textbf{A}\xrightarrow{} \textbf{W}$, the model trained with UDA baseline can achieve $94.50\%$ accuracy on clean target data; however, the model robustness against adversarial examples is $0.00\%$, as shown in the upper row of Figure\,\ref{fig:introduction}. Obviously, the robustness vulnerability of DNNs significantly hinders their applications in many real-world scenarios. Approaches to effectively improve the adversarial robustness of models produced by UDA is highly demanded.

In contrast to the intensive studies on UDA, few efforts have been devoted to exploring the robustness of UDA models. Recall that the most effective defense for adversarial attacks so far is adversarial training (AT), of which the core idea is to train the task model on adversarial examples that are online generated at each training epoch. However, it needs ground-truth labels to generate adversarial examples, which is inapplicable under the UDA scenario. A recent work~\cite{Awais2021ICCV} proposed to utilize an external adversarially pre-trained model as a teacher model to distill robustness knowledge during UDA process. However, its performance is limited by the teacher model’s perturbation budget and sensitive to the architecture of teacher model.



Recently, self-training~\cite{Xu_2019_ICCV_ST, Yang2022ST3D}, which trains the model on unlabeled target data in a supervised manner by utilizing pseudo labels generated from a pre-trained source model, has become popular as means of learning potential training signals in target domain for UDA. The insights of pseudo label generation in self-training provide us an access for directly injecting AT in UDA, where we can generate adversarial examples of target data with pseudo labels. However, we find that naive self-training does not work well in robust UDA training as shown in the middle row of Figure\,\ref{fig:introduction}. Although the model robustness can be improved, the clean accuracy decreased dramatically due to the inevitable noisy labels generated from source model and the odds between clean and robustness accuracy native in AT~\cite{Zhang2019tradeoff}.


In this paper, we propose SRoUDA, redesigning the self-training pipeline, for improving adversarial robustness of UDA models. First, in source model pre-training, we apply the UDA technique on source and target data with a developed random mask augmentation (RMA), to overcome the domain bias and initialize more proper pseudo labels for target data. Second, for target model training, we directly inject AT to train the target model using the adversarial examples generated from pseudo-labeled target data. The main challenge in this phase is how to improve pseudo label quality. Inspired by~\cite{Wang_meta_kdd, zhang2022noilin}, instead of using arbitrary pseudo labels for target model training, we form the pseudo label generation as an optimization problem and employ meta-learning with a designed meta-objective: the best pseudo labels should make the target model the best. Thus we propose a meta step in this stage where the source model is progressively updated by learning from the feedback of how the target network performs. In our implementation, the feedback signal is the performance of the target model on the labeled source data. It can reflect how much and how well the target model learns from the source model and brings alignment of source data with adversarial examples of target data, which consequently improves the target model’s performance. Specifically, the target and source models are trained alternatively: the target model learns robust knowledge from the pseudo-labeled target data, and then the source model is fine-tuned by the target model loss on the labeled source dataset.

We perform extensive experiments on various benchmarks and demonstrate the effectiveness of our approach, where the model robustness is improved by a large percentage ($0.00\% \xrightarrow{} 84.68\%$ in the bottom row of Figure\,\ref{fig:introduction}). Besides, the proposed method outperforms the state-of-the-art approach~\cite{Awais2021ICCV} by a notable margin on all evaluated settings. It is also noteworthy that our approach can even achieve a higher clean accuracy than the UDA baseline in several tasks.




\section{Related works}

\subsection{Unsupervised domain adaptation}

Unsupervised domain adaptation aims to transfer the knowledge learned from a labeled source data to an unlabeled target data. Conventional UDA approaches explore domain-invariant information across source and target data by minimizing the learned distribution discrepancy. Long~\textit{et~al}\mbox{.}~proposed DAN~\cite{long2015ICML} and JAN~\cite{long2017ICML} to minimize the feature discrepancy using Maximum Mean Discrepancy (MMD). More recently, the emergence of GANs has brought new inspiration to the field of domain adaptation, the DANN~\cite{Ganin2017ACVPR}, CDAN~\cite{Long2018NEURIPS}, and MCD~\cite{Saito2018CVPR} are proposed where a discriminator is equipped to force more discriminative domain-invariant feature generation. Despite the effectiveness of these methods, a drawback emerges that the potentially meaningful training information from the target domain are under-utilized. Thus, another line of works explored self-training scheme to generate pseudo labels for target domain and then re-train the model by pseudo-labeled target data. To improve the quality of pseudo labels, many efforts are devoted to reduce label noise by utilizing progressive generation strategy~\cite{Xu_2019_ICCV_ST}, curriculum learning~\cite{Choi2019PseudoLabelingCF}, and voting scheme~\cite{Yang2022ST3D,Fan_2022_CVPR}.

\begin{figure*}[t]
\begin{center}
\includegraphics[width=1.0\linewidth]{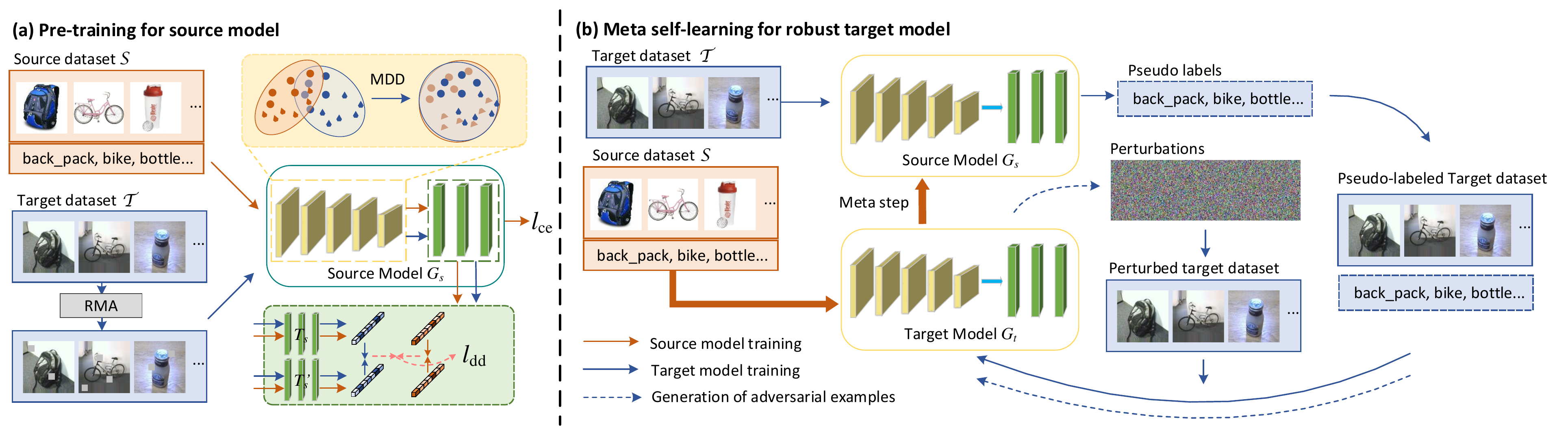}
\end{center}
\caption{An overview of our SRoUDA pipeline, which consists of two phases: (a) Pre-train the source model $G_s$ with RMA in target domain to mitigate the domain bias. (b) Train the target model on adversarial examples generated by pseudo-labeled target data, while fine-tuning the source model by employing a meta-step which is the feedback of the target model's performance on source labeled data.}
\label{fig:method}
\end{figure*}

\subsection{Adversarial robustness}
Since Szegedy~\textit{et~al}\mbox{.}~\cite{Szegedy} first reported that imperceptible perturbations can easily fool deep models, the adversarial vulnerability of deep models has gained increasing concerns. The work of~\cite{Goodfellow2015} designed a Fast Gradient Sign Method (FGSM) to produce strong adversarial examples based on the investigation of CNN linear nature. Madry~\textit{et~al}\mbox{.}~\cite{madry2018towards} further proposed Projected Gradient Descent (PGD) attack by changing the one-step perturbation generation in FGSM into iterative perturbation generation, and has become one of the most classic adversarial attacks. In response to the threat of adversarial examples, AT has been developed as a paradigm to train robust models. It is formed as a min-max game between adversarial example generation and model training. The work~\cite{madry2018towards} first formulated AT process, where they used PGD to generate adversarial examples and trained the model on these adversarial examples. Various modifications are developed to improve robustness accuracy of AT, with changes in the adversarial examples generation procedure~\cite{tramer2017ensemble,58kannan2018adversarial,zheng2020efficient}, model parameter updating~\cite{Hwang2021SWA} and feature adaption~\cite{Zhang2019tradeoff,wang2021adaptive}. However, AT requires labels and therefore not applicable under UDA settings.

\subsection{Adversarial robustness of UDA models}
In contrast to intensive studies on improving accuracy of UDA models, few efforts have been made to explore the adversarial robustness for UDA. Meanwhile, the main challenges for incorporating AT in UDA is the missing label information in target domain while AT needs ground-truth labels for generating adversarial examples. To address this issue, existing methods either skip the AT~\cite{Awais2021ICCV} or use self-supervised methods~\cite{lo2022exploring} to generate adversarial examples. For instance, Awais~\textit{et~al}\mbox{.}~\cite{Awais2021ICCV} directly explored the robustness transfer in UDA process instead of using the AT, and proposed to use an external pre-trained robust model for robust feature distillation during UDA process. Despite its effectiveness, its performance is limited by the teacher model’s perturbation budget and sensitive to the architecture of teacher model. On the other hand, Lo~\textit{et~al}\mbox{.}~\cite{lo2022exploring} proposed to use a self-supervised adversarial example generation for injection of AT into UDA. Sadly, such adversarial example generation cannot guarantee the inner-maximization in AT, thus leading to unsatisfied model robustness. Another naive self-training based method was proposed in \cite{Yang2021iccv} for image segmentation. Likewise, such simple pseudo label generation is not correct enough and might even bring an unsatisfied robustness, as we stated in previous section.

\section{Preliminaries}

\noindent\textbf{Problem setting.}
We set our problem under UDA scenario, where we have accessed to a labeled source dataset ${\mathcal{S}}=\{(x^{s}_{i},y^{s}_{i})\}^{n}_{i=1}$ and an unlabeled target dataset ${\mathcal{T}}=\{(x^{t}_{i})\}^{m}_{i=1}$. Considering a classification model composed of a feature extractor, $F: x \xrightarrow{} f$, where $f$ is the feature representation of original input $x$, and a classification layer, $T: f \xrightarrow{} z$, where $z$ is the logit output, and $z \in \mathbb{R}^{C}$, $C$ denotes the number of classes. The goal of UDA is to produce a target model with high accuracy on the unlabeled target dataset. Here we go further to achieve not only accurate but also robust model on target domain. We first revisit the conventional UDA and AT algorithms.

\noindent\textbf{UDA.} A typical UDA model is optimized by a source error plus a discrepancy metric between the target and the source as follows:
\begin{align}\label{uda}
	\min_{F,T} \mathcal{L}_{uda}=\ell_{ce}(T(F(x^s)), y^s) + \ell_{da}(F(x^s), F(x^t)),
	\end{align}
where $\ell_{ce}$ denotes the cross-entropy loss for classification task, $\ell_{da}$ is the domain adaptation loss defined by different UDA approaches.   

\noindent\textbf{AT.} An adversarial example is an perturbed image which is obtained by adding a perturbation $\delta$ over original data $x$:
\begin{equation}
\hat{x}=x+\delta, ~~~~\text{s.t.}~~~ \hat{x} \in \mathcal{B}(x),
\end{equation}
where $\mathcal{B}(x)$ denotes the $\ell_{p}$-norm ball centered at $x$ with radius $\epsilon$, i.e., $\mathcal{B} (x)=\{\hat{x}:\|\hat{x}-x\|_{p} \leq \epsilon\}$. Correspondingly, the AT process directly takes adversarial examples as training data and has the following objective:
	\begin{align}\label{eq1}
		\min _{F, T} \mathcal{L}_{at}= \ell_{ce}(T(F(\hat{x})), y).
	\end{align}
	
Typically, the adversarial examples in AT are generated using PGD attack in an iterative way as follows:
\begin{align}\label{eq:at}
	\hat{x}_{k+1} & = \Pi_{\mathcal{B}(x)}(\hat{x}_{k}+\alpha\operatorname{sign}(\nabla_{x} \ell_{ce}(T(F({x})), y))) ,
\end{align}
where $\hat{x}_{k}$ is initialized as the clean input $x$, and the final adversarial example $\hat{x} = \hat{x}_{k_{max}}$, where $k_{max}$ is the maximum number of iterations; and $\Pi$ refers to the projection operation for projecting the adversarial examples back to the norm-ball.

\section{Methodology}
We now introduce our SRoUDA for training a not accurate but also robust model under the UDA settings. An overview of our framework is shown in Figure\,\ref{fig:method}. Starting from pre-training a source model $G_s$ by applying UDA baseline on source labeled data $\mathcal{S}$ and taraget unlabeled data $\mathcal{T}$ with random masked augmentation in Figure\,\ref{fig:method} (a), our SRoUDA alternates between adversarial target model training on pseudo-labeled target data and fine-tuning source model by a meta step in Figure\,\ref{fig:method} (b).




\subsection{Source model pre-training}\label{sec:pre-train}
The source model $G_s$ learns how to perform classification task on source labeled data $\mathcal{S}$ and is further adopted to produce pseudo labels $\Bar{y}^t$ for the unlabeled target data $\mathcal{T}$. To better initialize pseudo labels for the second phase, we consider the design of source model pre-training from two aspects: 1) reduce the learning bias from the source domain $\mathcal{S}$; 2) improve the model generalization for target domain $\mathcal{T}$. To this end, instead of only training on source data, we adopt UDA baseline to utilize target data for reducing the model learning bias from source domain. Moreover, to improve the generalization of pre-trained model, inspired by the masked autoencoders (MAE)~\cite{He_2022_CVPR_MAE}, we propose a very simple yet effective augmentation strategy, i.e., RMA, to facilitate the object-aware recognition. 

\vspace{3pt}
\noindent\textbf{Random masked augmentation.} Given a target image $x^t \in \mathbb{R}^{h\times w \times c}$, we first divide the target image into non-overlapping patches $z\in \mathbb{R}^{1\times \frac{hw}{2^2}}$ of size $2 \times 2$ . RMA then samples a subset of the patches with a uniform distribution and masks the remaining patches, which can be presented as:
\begin{align}\label{eq:rma}
z^+_i = \begin{cases}
z_i, & i\in R \\
0,& i\notin R \\
\end{cases}~,
\end{align}
where $R\sim U(1, \frac{hw}{2^2})$. Different from the image reconstruction task in MAE, where they set the sampling ratio as $75\%$ for learning content reconstruction, here we set the sampling ratio as $25\%$ to cover most of the image contents for data augmentation. The final augmented image $x^{t+}$ is formed by grouping $z^+_i$ back into an image. In the pre-training process, the target data is the combination of the original and augmented data.

\vspace{3pt}
\noindent\textbf{Pre-training.} Before the interaction between source and target models, we first pre-train the source model for a better initialization of pseudo labels for target data. Here we adopt the basic UDA approach Margin Disparity Discrepancy (MDD) to pre-train a clean UDA model. Specifically, MDD first employs an auxiliary classifier $T_s'$ to evaluate the disparity discrepancy as 
\begin{equation}
    \begin{split}\label{eq:dd}
   \ell_{dd} (x^s, x^t)= &\ell_{ce}(T_s'(F_s(x^t)), T_s(F_s(x^t))) \\
   - &\gamma \ell_{ce}(T_s'(F_s(x^s)), T_s(F_s(x^s))),
\end{split}
\end{equation}
where $\gamma$ is the margin factor and is set to be $4$. While auxiliary classifier $T_s'$ is trained to maximize the disparity discrepancy loss $\ell_{dd}$ in Eq.\,\eqref{eq:dd}, the source model is updated as:
\begin{align}\label{eq:pretrain}
	\min_{F_s, T_s} \mathcal{L}_{MDD}=\ell_{ce}(T_s(F_s(x^s)), y^s) + \eta \ell_{dd}(x^s, x^t),
\end{align}
where $\eta$ is a regularization factor which can be set as $0.1$ according to~\cite{Zhang2019ICML}. Note that it is not compulsory to choose MDD as the UDA baseline for pre-training here. Please refer to the experimental section for more comparisons of using different UDA baselines in source model pre-training.

	\begin{algorithm}[!t]
		\SetKwFunction{isOddNumber}{isOddNumber}
		\SetKwInOut{KwIn}{Input}
		\SetKwInOut{KwOut}{Output}
		\KwIn{Source domain labeled data ${\mathcal{S}}=\{(x^{s}_{i},y^{s}_{i})\}$, and target domain unlabeled dataset ${\mathcal{T}}=\{(x^{t}_{i})\}$. Source model $G_s$, target model $G_t$, batch size $B$, learning rate $lr$, and iteration number $N$.
		}
		\KwOut{The adversarial robust model for target domain $G_t$.}
	    Pre-train the source model $G_s$ as detailed in Eqs.\eqref{eq:rma}-\eqref{eq:pretrain}. 
	    
	    Initialize the target model $G_t$ by copying parameters from the source model $G_s$.\\
		\For{epoch = 1 ... $N$ }{
				Sampling a random mini-batch of training samples $x^t \xleftarrow{} \{x^{t}_{i}\}_{i=1}^{B}$,  $x^s \xleftarrow{} \{x^{s}_{i},  y^{s}_{i}\}_{i=1}^{B}$;
				
				Generating pesudo labels $\Bar{y}^t$ for target data $x^t$;
				
				Generating adversarial examples of target data $\hat{x}^t$ based on the pesudo labels by Eq.\,\eqref{eq:at-};
				
				Update $G_t$ by optimizing adversarial training loss in Eq.\,\eqref{eq:att};
				
				Utilize the current target model $G_t$ to compute meta loss on source data $x^s$ in Eq.\,\eqref{eq:meta-loss};
				
				Update $G_s$ by optimizing the meta loss;
				
		} 
		\caption{Meta self-training for robust unsupervised domain adaptation (SRoUDA)}
		\label{al: trianing}
		
	\end{algorithm}

\subsection{Meta self-training for robust target model}
With the pre-trained source model, the next phase is to generate pseudo labels for the unlabeled target data and then the target model can be trained on the adversarial examples generated by the pseudo labeled target data, which can be presented as:
\begin{align}\label{eq:overall}
\min _{G_t} \mathcal{L} = \ell_{ce}(G_t(\hat{x}^t), \Bar{y}^t),
\end{align}
where $\hat{x}^t$ is the adversarial example of target data and $\Bar{y}^t = G_s(x^t)$ is the pseudo label for $x^t$. It is obvious that the optimal $G_t$ heavily depends on the source model $G_s$ via the pseudo labels. Considering the noisy labels are inevitable in the initial pseudo labels produced from pre-trained model, we need to progressively fine-tune the source model $G_s$ for better pseudo labels that can make the target model best. We thus consider the pseudo label generation as an optimization problem, and transform Eq.\,\eqref{eq:overall} into:
\begin{align}\label{eq:labelop}
\min _{G_t, G_s} \mathcal{L} = \ell_{ce}(G_t(\hat{x}^t), G_s(x^t)).
\end{align}

\begin{table*}[t]
    \centering
    \caption{Comparison of clean and robustness accuracy $\%$ of different UDA models produced by different methods on \textbf{Office-31} dataset. Note that $\dag$ denotes the results are directly copied from the original paper. We only show the average accuracy of RFA since they did not include detailed number in~\cite{Awais2021ICCV}.}\label{tab: office}
    \resizebox{\textwidth}{!}{
    \begin{tabular}{c|cc|cc|cc|cc|cc|cc|cc}
       \toprule
     
        \multirow{2}{*}{Methods}&\multicolumn{2}{|c|}{\textbf{A} $\xrightarrow{}$ \textbf{W}}&\multicolumn{2}{|c|}{\textbf{D} $\xrightarrow{}$ \textbf{W}}&\multicolumn{2}{|c|}{\textbf{A} $\xrightarrow{}$ \textbf{D}}&\multicolumn{2}{|c|}{\textbf{D} $\xrightarrow{}$ \textbf{A}}&\multicolumn{2}{|c|}{\textbf{W} $\xrightarrow{}$ \textbf{D}}&\multicolumn{2}{|c|}{\textbf{W} $\xrightarrow{}$ \textbf{A}}&\multicolumn{2}{|c}{\textbf{Avg.}}\\
         
        \cmidrule{2-15}
        &Clean&Rob. &Clean&Rob. &Clean&Rob. &Clean&Rob. &Clean&Rob. &Clean&Rob. &Clean&Rob.\\
        \midrule
        Baseline (UDA) $\dag$ &94.50 &0.00 &98.40 &0.00 &93.50 &0.00 &74.60  &0.65 &100.00 &0.00 &72.20 &1.93 &88.90 &0.43 \\
        \midrule

        Source only &28.23 &6.45 &45.16 &26.61 &16.13 &6.45 &9.46  &5.59 &62.90 &25.81 &12.69 &5.81 &29.10 &12.79 \\
        \midrule
        

        AT+UDA &74.09 &34.47 &91.19 &70.57 &73.49 &19.28 &39.65 &24.46 &98.59 &66.87 &55.31 &35.39 &72.02 &42.21 \\ \midrule
        
        UDA+AT&75.81 &62.90 &91.13 &53.23 &83.87 &43.55 &64.52 &53.33 &95.16 &51.61 &64.52 &53.12 &79.78 &53.21 \\ \midrule
        
        RFA~\cite{Awais2021ICCV} $\dag$ &- &- &- &- &- &- &- &-&- &-&- &- &84.21 &74.31 \\ \midrule
            
        SRoUDA (Ours) &95.97 &84.68 &96.77 &83.87 &91.94 &85.48 &72.47 &57.20 &100.00 &88.71 &67.10 &57.42 &87.27 &75.79 \\

        \bottomrule
    \end{tabular}}
\end{table*}

\begin{table*}[t]
    \centering
	\caption{Comparison of clean and robustness accuracy $\%$ of different UDA models produced by different methods on \textbf{Digits} and \textbf{CIFAR} datasets.} \label{tab: atdataset}
 	\resizebox{\textwidth}{!}{
    \begin{tabular}{c|cc|cc|cc|cc|cc|cc|cc}
       \toprule
     
        \multirow{2}{*}{Methods}&\multicolumn{2}{|c|}{\textbf{M} $\xrightarrow{}$ \textbf{U}}&\multicolumn{2}{|c|}{\textbf{U}$\xrightarrow{}$ \textbf{M}}&\multicolumn{2}{|c|}{\textbf{S}$\xrightarrow{}$ \textbf{M}}&\multicolumn{2}{|c|}{\textbf{Avg.}} &\multicolumn{2}{|c|}{\textbf{CIFAR}$\xrightarrow{}$\textbf{STL}}&\multicolumn{2}{|c|}{\textbf{STL}$\xrightarrow{}$\textbf{CIFAR}}&\multicolumn{2}{|c}{\textbf{Avg.}}\\
         
        \cmidrule{2-15}
        &Clean&Rob. &Clean&Rob. &Clean&Rob. &Clean&Rob. &Clean&Rob. &Clean&Rob.&Clean&Rob.\\
        \midrule
        Baseline(UDA)&95.60 &12.31 &97.36 &35.23 &89.20 &41.19 &94.05 &29.58 &68.82 &9.28 &59.16 &13.29 &63.99 &11.29\\
        \midrule
        source only&76.13 &57.90 &54.63 &48.59 &22.63 &15.14 &51.13 &40.54 &42.71 &24.08 &35.53 &21.41 &39.12 &22.75 \\
        \midrule
        AT+UDA&93.42 &69.76 &96.35 &74.83 & 73.83 &67.55 &87.87 &70.71  &50.33 &28.51 &33.37 &18.80 &41.85 &23.66\\
        \midrule
        
        UDA+AT&92.43 &83.26 &97.85 &93.56 &65.09 &63.64 &85.12 &80.15  &58.64 &36.86 &40.72 &25.53 &49.68 &31.20\\
        \midrule

        SRoUDA (Ours) &95.02&87.59&98.50&96.44 &88.72 &87.16 &94.08 &90.40 &50.75 &31.57 &62.04 &38.50 &56.40 &35.04\\

        \bottomrule

    \end{tabular}}
\end{table*}

The object in Eq.\,\eqref{eq:labelop} can be regarded as a joint learning of $G_t$ and $G_s$. Inspired by meta learning~\cite{Pham_2021_CVPR,Wang_meta_kdd}, we design a meta step for source model fine-tuning with a meta-objective that the best pseudo labels should improve the learning of target model, where the feedback signal is the performance of target model on source labeled data. Specifically, in each iteration, the target model $G_t$ is trained on the adversarial examples generated by the target data with pseudo labels that are produced by the source model. Then the source model $G_s$ is fine-tuned by learning from the feedback of the performance of target model on source dataset. By this way, the pseudo labels can be adjusted accordingly to further improve target model's performance. The training steps are as follows:

\textbf{Step 1.} Given fixed pre-trained source model $G_s$, the target model $G_t$ performs adversarial training on adversarial examples of target data that are generated by pseudo labels produced by $G_s$. In each iteration, the adversarial examples of target data are first generated by changing Eq.\,\eqref{eq:at} to:
\begin{align}\label{eq:at-}
	\hat{x}^t_{k+1} & = \Pi_{\mathcal{B}}(\hat{x}^t_{k}+\alpha\operatorname{sign}(\nabla_{x^t} \ell_{ce}(G_t(x^t), G_s(x^t))) ,
\end{align}
where $\hat{x}^t= \hat{x}^t_{k_{max}}$. Typically, $k_{max}$ is set as 10 during AT process. Then the target model is further trained on the generated adversarial examples as follows: 
\begin{align}\label{eq:att}
\min _{G_t} \mathcal{L}_{at} = \ell_{ce}(G_t(\hat{x}^t), G_s(x^t)).
\end{align}

\textbf{Step 2.} Next, to improve the quality of pseudo labels, we apply a meta-step to utilize the performance of $G_t$ on source data for fine-tuning the source network. The meta loss is computed as:
\begin{align}\label{eq:meta-loss}
\mathcal{L}_{meta} = \ell_{ce}(G_t(x^s), y_s).
\end{align}
 
The source model $G_s$ is then fine-tuned by the gradient descent based on $\mathcal{L}_{meta}$. The algorithm of our SRoUDA is summarized in Algorithm\,\ref{al: trianing}.

\section{Experiments}
\subsection{Experimental setup}
\noindent\textbf{Datasets.} We evaluate our method on the both main-stream UDA benchmark datasets and AT datasets: 1) \textbf{Office-31} dataset, which is a standard domain adaptation dataset with three domains: Amazon (\textbf{A}, 2,817 images), Webcam (\textbf{W}, 795 images), and DSLR (\textbf{D}, 498 images), It is imbalanced across domains. 
2) \textbf{Digits} dataset containing 3 different domains: MNIST (\textbf{M}), USPS (\textbf{U}), and SVHN (\textbf{S}). Note that the images in \textbf{M}, \textbf{U} are gray-scale, whereas the images in \textbf{S} are colored. 3) \textbf{CIFAR} and \textbf{STL} datasets. Both datasets contain 10 categories, of which the overlapping categories are 9 categories. We remove the different categories in the two datasets and changing the 10-category classification task into 9-category task.

\noindent\textbf{Compared methods.} We compare our SRoUDA with four baselines: (1) UDA baseline: UDA method without considering model robustness; (2) Source only: the model is adversarially trained on only source data; (3) UDA+AT: injecting AT into UDA process based on the naive self-training pipeline, where a source model is first pre-trained by UDA, and then the target model is adversarially trained on pseudo-labeled target data; (4) AT+UDA: the source data is first transferred into adversarial examples, and then perform UDA on adversarial source data and clean target data. We also compare with the state-of-the-art method RFA using
their reported results in ~\cite{Awais2021ICCV}.

\noindent\textbf{Implementation details.} We validate our proposed SRoUDA on three backbones for fair comparison with SOTAs. Specifically, we use ResNet-50 on \textbf{Office-31}, DTN on \textbf{Digits} dataset, and WideResNet-50-2 on \textbf{CIFAR} dataset. During the pre-training of source model, we adopt the training settings of the popular UDA codebase TLlib and train the source model for 20 epochs with a learning rate of $0.004$. For the following meta self-training stage, we iteratively update the source and target models. In the AT process, we set $k_{max}=10$, $\epsilon=8/255$ in adversarial example generation, the Adam optimizer with learning rate $0.0015$ is used to update the target model. We update the source model every epoch in this process. During both the pre-training and meta self-training processes, we also adopt the widely used data augmentation, including random flipping, and rotation for avoiding overfitting.

\begin{table}[t]
    \centering
    \caption{Robustness comparison against different adversarial attacks on \textbf{STL} $\xrightarrow{}$ \textbf{CIFAR-10} task.}\label{tab:diffattack}
    \fontsize{8}{9.5}\selectfont
    \begin{tabular}{c|ccccc}
       \toprule
     
        \multirow{2}{*}{Method}&\multicolumn{5}{|c}{\textbf{STL} $\xrightarrow{}$ \textbf{CIFAR-10}}\\
         
        \cmidrule{2-6}

        &Clean &FGSM &PGD-10 &PGD-20 &CW$_\infty$ \\
        \midrule
        
       Baseline &59.16 &32.07 &16.73  &13.29  &3.59 \\ \midrule
       Source only &36.89 &20.61  &13.47  &12.11  &12.23 \\ \midrule
       AT+UDA &33.37 &21.96 &19.28  &18.80  &12.56  \\ \midrule
       UDA+AT  &40.72  &40.15 &25.54 &25.53 &24.21 \\ \midrule
       SRoUDA (Ours) &\textbf{62.24}  &\textbf{41.47} &\textbf{38.82}  &\textbf{38.50}   &\textbf{37.23} \\ 
        
       \bottomrule
    \end{tabular}
\end{table}

\begin{table}[t]
    \centering
    \caption{Experimental results on component ablations of SRoUDA.} \label{tab:compoablation}
    \fontsize{9}{11}\selectfont
    \begin{tabular}{c|cc|cc}
       \toprule
     
        \multirow{2}{*}{Method}&\multicolumn{2}{|c|}{\textbf{W} $\xrightarrow{}$ \textbf{A}}&\multicolumn{2}{|c}{\textbf{D} $\xrightarrow{}$ \textbf{A}}\\
         
        \cmidrule{2-5}
        &Clean&Rob. &Clean&Rob.\\
        \midrule
        SRoUDA  w/o pre-train&50.75 &40.22 &40.43 &22.15 \\
        \midrule
        SRoUDA w/o meta-step &64.52 &53.12 &68.17 &54.84\\
        \midrule
        SRoUDA  w/o RMA &63.44 &51.40 &72.04 &52.90\\
        \midrule
        SRoUDA  &\textbf{67.10} &\textbf{57.42}  &\textbf{72.47}  &\textbf{57.20}\\
       \bottomrule
    \end{tabular}
\end{table}

\subsection{Main results}
\noindent\textbf{Overall results.} We present the comprehensively comparison with different methods in Table\,\ref{tab: office}-\ref{tab: atdataset}. Here we use MDD as the UDA baseline and PGD-20 attack to test the models' robustness. 

From Table\,\ref{tab: office}-\ref{tab: atdataset}, we can have the following observations. First, the proposed SRoUDA can effectively improve adversarial robustness of UDA models, which significantly outperforms the other UDA baselines. Specifically, we improve the performance of model robustness on all the datasets by a large margin, e.g., $0.43 \%$ to $75.79 \%$ on average for \textbf{Office-31}; $29.58 \%$ to $90.40 \%$ on average for \textbf{Digits}. Furthermore, when tested on small-scale datasets such as \textbf{Digits}, our method can achieve both near-optimal accuracy and robustness simultaneously on target domain, i.e., $98.50\%$ clean and $96.44\%$ robustness accuracy in $\textbf{U} \xrightarrow{} \textbf{M} $ task. These encouraging results validate that our SRoUDA can effectively enhance the adversarial robustness of UDA models and perform generally well on different domain adaptation tasks.

Second, compared to other schemes for improving UDA robustness, our method achieves the state-of-the-art model robustness without harming clean accuracy. Although the naive self-training UDA can improve the robustness, the clean accuracy is dramatically dropped, e.g., $88.90 \%$ to $79.78 \%$ on average for \textbf{Office-31}; and $94.05 \%$ to $85.12 \%$ on average for \textbf{Digits}. This risk comes from both the pseudo label noises and the odds between accuracy and robustness native in AT. On the contrary, our method can even improve the clean accuracy on some tasks, e.g., for \textbf{Office-31} dataset, SRoUDA improves clean accuracy over the UDA baseline on $\textbf{A} \xrightarrow{} \textbf{W}$ ($94.50 \%$ to $95.97 \%$); for \textbf{Digits} dataset, our method improves $\textbf{U} \xrightarrow{} \textbf{M}$ ($97.36 \%$ to $98.50 \%$). This benefits from the meta-step for fine-tuning source model in target model training with the objective to perform well on source data, which inherently aligns the source data with the adversarial examples of target data during target model training.


We also test the model robustness against different adversarial attacks on $\textbf{STL} \xrightarrow{} \textbf{CIFAR-10}$ task in Table\,\ref{tab:diffattack}. Here, the FGSM, PGD-10, PGD-20, CW$_\infty$ attacks are used to evaluate the model robustness. It can be observed that the model produced by our SRoUDA can achieve higher robustness accuracy on all attack settings, demonstrating that our method can produce robust models against multiple attacks, which is crucial for practical employment of DNNs.

\begin{table}[t]
    \centering
    \caption{Comparison of using different UDA baselines in source model pre-training on \textbf{W} $\xrightarrow{} $ \textbf{D} and \textbf{D} $\xrightarrow{} $ \textbf{A} tasks.}\label{tab:udaablation}
    \fontsize{9}{11}\selectfont
    \begin{tabular}{c|cc|cc}
       \toprule
     
        \multirow{2}{*}{UDA baselines}&\multicolumn{2}{|c|}{\textbf{W} $\xrightarrow{} $ \textbf{D}}&\multicolumn{2}{|c}{\textbf{D} $\xrightarrow{} $ \textbf{A}}\\
         
        \cmidrule{2-5}

        &Clean&Rob. &Clean&Rob.  \\
        \midrule
        
       DAN~ &91.94 &82.26 &59.78 &43.23 \\ \midrule
       
       DANN~ &98.39 &83.87 &68.60 &49.25 \\ \midrule
       
       JAN~ &96.77 &87.10 &62.58 &48.82 \\ \midrule
       
       CDAN~  &98.39  &72.58 &70.97  &51.61 \\ \midrule
       
       MDD~ &\textbf{100.00}   &\textbf{88.71} &\textbf{72.47}  &\textbf{57.20} \\
        
       \bottomrule
    \end{tabular}
\end{table}

\begin{figure}[t]
\begin{center}
\includegraphics[width=1.0\linewidth]{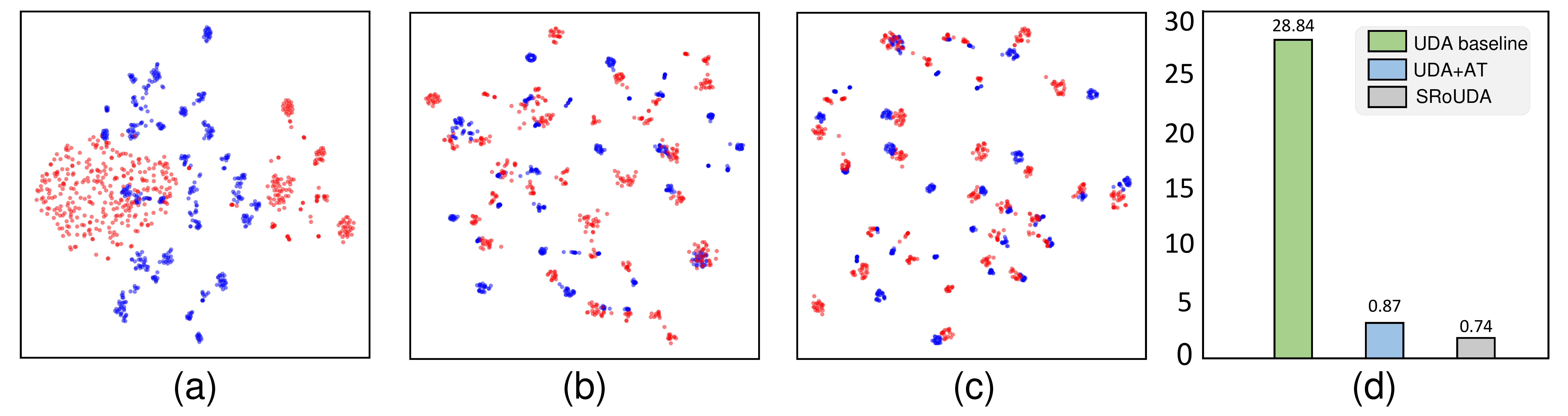}
\end{center}
\vspace{-5pt}
\caption{ (a)-(c): The t-SNE visualization of extracted features from target models trained with UDA baseline, UDA+AT with naive self-training, and our method on \textbf{W} $\xrightarrow{}$ \textbf{D} task, respectively. The \textcolor{blue}{blue} dots denote clean target data and \textcolor{red}{red} dots denote the adversarial examples of target data. (d): The mean $L_2$-norm distance of clean and adversarial examples in feature space.}
\label{fig:t-sne}
\vspace{-10pt}
\end{figure}


\begin{figure*}[h]
\begin{center}
\includegraphics[width=1.0\linewidth]{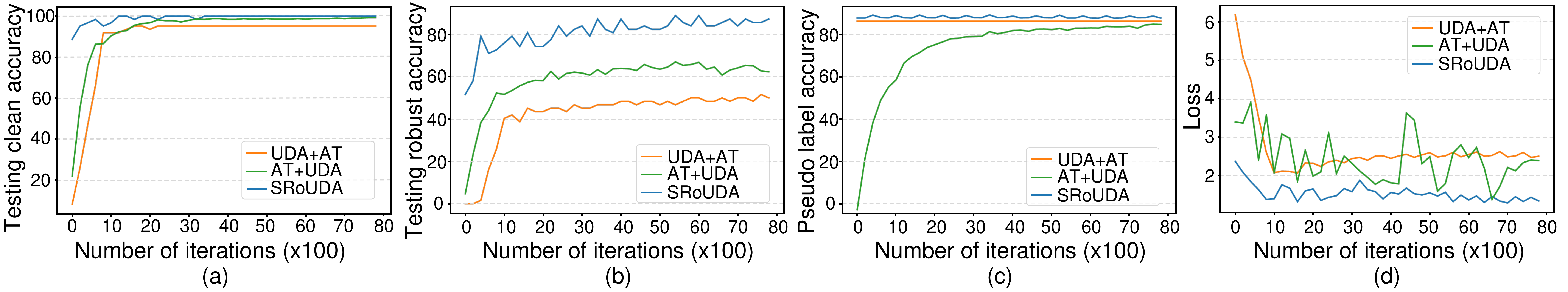}
\end{center}
\caption{The training curve of target model using different training schemes. (a) Testing clean accuracy convergence, (b) testing robust accuracy convergence, (c) pseudo label accuracy convergence, and (d) training loss convergence. Note that for SRoUDA, we show the training convergence of target model in the second training phase.}
\label{fig:convergence}
\vspace{-8pt}
\end{figure*}

\subsection{Analysis}
We analyze our SRoUDA from four perspectives: (1) component ablations of the SRoUDA pipeline, (2) sensitivity to the UDA techniques in the pre-training, (3) feature space analysis, and (4) training convergence.


\subsubsection{Component ablations} We examine the effectiveness of individual components in our SRoUDA, and conduct the comparison on \textbf{Office-31} $\textbf{W} \xrightarrow{} \textbf{A}$ and $\textbf{D} \xrightarrow{} \textbf{A}$ tasks following the experimental settings described previously. The results are shown in Table\,\ref{tab:compoablation}. Specifically, we testify 
(1) SRoUDA w/o pre-train: removing the source model pre-training and randomly initialize the source model parameters.We can clearly see that without pre-training on the source model, both the clean and robustness accuracy drop by a large margin because of the misleading initialized pseudo labels, which indicates the pre-training of source model is crucial for self-training pipeline. 
(2) SRoUDA w/o meta-step: removing the meta learning step and use fixed pseudo labels for training robust target model. Using pre-trained source model but without fine-tuning it, SRoUDA w/o meta-step performs better than SRoUDA w/o pre-train but still suffers from over $10$\% degradation on both clean and adversarial accuracy compared with SRoUDA. 
(3) SRoUDA w/o RMA: removing the data augmentations. Without data augmentation, our SRoUDA suffers degradation both the clean and robustness accuracy on $\textbf{W} \xrightarrow{} \textbf{A}$ and $\textbf{D} \xrightarrow{} \textbf{A}$ tasks. In general, the full SRoUDA pipeline can achieve both the highest clean and robustness accuracy on target domain.




\subsubsection{Different UDA techniques} Besides using MDD as the UDA baseline, we investigate our SRoUDA pipeline with different UDA techniques. Here we use DAN, DANN, JAN, and CDAN to replace the MDD in source model pre-training of our SRoUDA pipeline on $\textbf{W} \xrightarrow{} \textbf{D}$ and $\textbf{D} \xrightarrow{} \textbf{A}$ tasks. The results are reported in Table\,\ref{tab:udaablation}. Both the clean and robustness accuracy vary as the change of UDA baselines in source model pre-training, which once again indicates that the model pre-training plays a crucial role in self-training pipelines. Based on the performance of different UDA techniques in natural training, we can see that better UDA baseline can help produce more robust target model since the pseudo labels are more reliable.

\subsubsection{Feature space analysis} We also visualize the feature generalization in the target model trained by different methods using t-SNE embeddings in Figure\,\ref{fig:t-sne} (a)-(c). The features are extracted from the last convolution layer of the target model. As we can see, the adversarial examples reside in a large region that are hard to distinguish in natural trained UDA model (Figure\,\ref{fig:t-sne} (a)). By applying naive self-training scheme on UDA, the adversarial examples are discriminated better by incorporating adversarial training but are not well aligned with the clean data due to the bias of pseudo labels. In contrast, our method is evidently better and the categories are well discriminated. We further compute the mean $L_2$-norm distance between the clean target data and its corresponding adversarial examples in the feature space as $\|F_t(x^t)-F_t(\hat{x}^t)\|_2$. The results are given in Figure\,\ref{fig:t-sne} (d). As we can observe, the UDA baseline has the largest distance as it does not consider the model robustness. Incorporating AT can greatly minimize the distance, while our SRoUDA achieves the lowest distance.

\subsubsection{Convergence} We testify the convergence of UDA+AT, AT+UDA and SRoUDA with the testing clean accuracy, testing robustness accuracy, pseudo label accuracy, and loss function on task \textbf{W} $\xrightarrow{}$ \textbf{D} shown in Figure\,\ref{fig:convergence}. With a sophisticated pre-trained source model, SRoUDA enjoys a faster convergence than UDA+AT and AT+UDA by taking the source model as initialization. For UDA+AT, the pseudo label generation is fixed thus the performance of clean and robustness accuracy is limited. For AT+UDA, as minimizing the discrepancy between source adversarial examples and clean data, the clean accuracy on target data can be improved. However, due to the inherent difference between source adversarial examples and target adversarial examples, the robustness accuracy on target data is less satisfied. In contrast, by equipping with a meta step to progressively adjust the pseudo labels, the clean and robustness accuracy of SRoUDA can be improved steadily as the training goes deeper shown in Figure\,\ref{fig:convergence} (a) and (b). Although the pseudo label accuracy does not improve that much during the fine-tuning of source model in Figure\,\ref{fig:convergence} (c), but the loss continues to be lower in Figure\,\ref{fig:convergence} (d). This phenomenon verifies the view of Zhang et al~\cite{zhang2022noilin}, adversarial robustness can be enhanced by noisy label injection, they are optimized towards a better robust target model training.

\section{Conclusion}
In this paper, we tackle the problem of model robustness of unsupervised domain adaption models. We have presented SRoUDA, a redesigned self-training pipeline for robust unsupervised domain adaptation. SRoUDA involves a pre-training stage with a simple but effective random masked augmentation for source model, and a meta self-training stage for alternatively training robust target model and fine-tuning source model. Extensive experiments demonstrate that SRoUDA can effectively improve the robustness of UDA models by a large margin over baselines under various defense settings.

\section{Acknowledgements}
This work was partly supported by the National Natural Science Foundation of China under Grant Nos. 62202104, 62102422, 62072109 and U1804263; the Ministry of Science and Technology, Taiwan, under Grant MOST 111-2628-E-155-003-MY3; and Youth Foundation of Fujian Province, P.R.China, under Grant No. 2021J05129.

\bibliography{Formatting-Instructions-LaTeX-2023}

\end{document}